\documentclass[10pt,twocolumn,letterpaper]{article}

\usepackage[pagenumbers]{cvpr}

\usepackage{graphicx}
\usepackage{amsmath}
\usepackage{amssymb}
\usepackage{booktabs}

\usepackage{multirow}
\usepackage{algorithm}
\usepackage{algorithmic}
\usepackage[table]{xcolor}
\usepackage{pifont}
\usepackage{booktabs}

\DeclareMathOperator*{\argmin}{arg\,min}
\definecolor{baselinecolor}{HTML}{EFEFEF}
\usepackage[pagebackref,breaklinks,colorlinks]{hyperref}

\usepackage[capitalize]{cleveref}
\crefname{section}{Sec.}{Secs.}
\Crefname{section}{Section}{Sections}
\Crefname{table}{Table}{Tables}
\crefname{table}{Tab.}{Tabs.}

\def\cvprPaperID{****} % *** Enter the CVPR Paper ID here
\def\confName{CVPR}
\def\confYear{2023}

\begin{document}

\title{Domain Adaptive Scene Text Detection via Subcategorization}

\author{Zichen Tian\textsuperscript{1}, Chuhui Xue\textsuperscript{2}, Jingyi Zhang\textsuperscript{1}, Shijian Lu\textsuperscript{1}\\
\textsuperscript{1}Nanyang Technological University, \textsuperscript{2}ByteDance Inc.\\
{\tt\small {zichen.tian,jingyi.zhang}@ntu.edu.sg, xuec0003@e.ntu.edu.sg}
}
\maketitle

\begin{abstract}
Most existing scene text detectors require large-scale training data which cannot scale well due to two major factors: 1) scene text images often have domain-specific distributions; 2) collecting large-scale annotated scene text images is laborious. We study domain adaptive scene text detection, a largely neglected yet very meaningful task that aims for optimal transfer of labelled scene text images while handling unlabelled images in various new domains. Specifically, we design SCAST, a subcategory-aware self-training technique that mitigates the network overfitting and noisy pseudo labels in domain adaptive scene text detection effectively. SCAST consists of two novel designs. For labelled source data, it introduces pseudo subcategories for both foreground texts and background stuff which helps train more generalizable source models with multi-class detection objectives. For unlabelled target data, it mitigates the network overfitting by co-regularizing the binary and subcategory classifiers trained in the source domain. Extensive experiments show that SCAST achieves superior detection performance consistently across multiple public benchmarks, and it also generalizes well to other domain adaptive detection tasks such as vehicle detection. Codes are available at \url{https://github.com/doem97/SCAST}.
\end{abstract}

\section{Introduction}
\begin{figure}[t]
    \centering
    \includegraphics[width=0.95\linewidth]{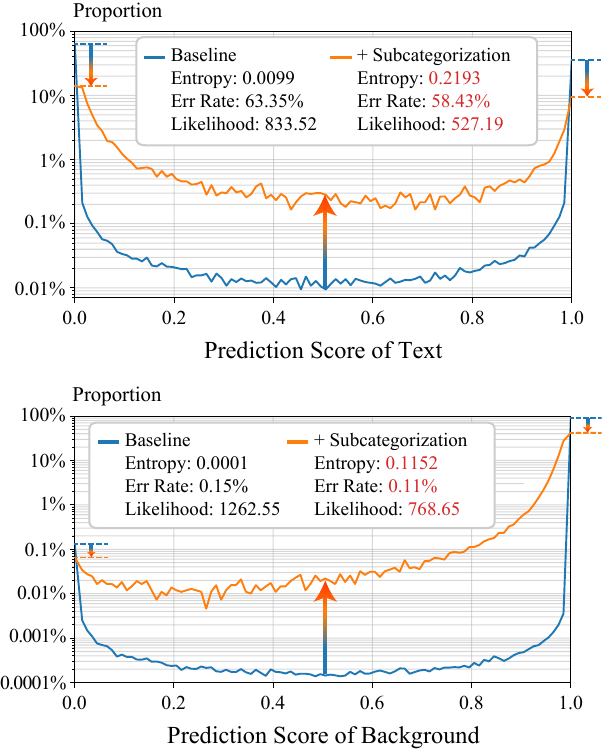}
    \caption{The proposed subcategorization technique mitigates the overfitting in domain adaptive scene text detection: Existing domain adaptive scene text detector suffers from clear overfitting with over-confident predictions with most prediction probabilities lying around $0$ or $1$ (blue curves). The proposed subcategorization mitigates the overfitting effectively (orange curves), leading to higher entropy and lower pseudo-labeling error. The studies adopt scene text detector EAST~\cite{zhou2017east} over domain adaptive scene text detection task SynthText $\to$ IC15.\vspace{-0.7em}}
    \label{fig:distrib}
\end{figure}

Scene text detection has been studied intensely for years, largely due to its wide applications in many real-world tasks such as scene understanding~\cite{biten2019scene, singh2019towards}, image retrieval~\cite{wang2021scene, gomez2018single}, autonomous indoor and outdoor navigation~\cite{long2021scene}, etc. With the recent advances of deep neural networks, it has achieved significant progress under the presence of large-scale annotated training images~\cite{liao2017textboxes, tian2016detecting, zhou2017east, liao2018rotation, lyu2018mask, zhang2019look, wang2019shape, dai2021progressive, liao2020real, zhan2019ga}. However, collecting large-scale annotated scene text images is laborious which has become a bottleneck while facing various scene text detection tasks that often have different camera setups, environmental parameters, etc.

Unsupervised domain adaptation (UDA), which aims to learn from a labeled \textit{source domain} for a well-performing model in an unlabelled \textit{target domain}, has achieved great success in different computer vision tasks such as semantic segmentation~\cite{grandvalet2005semi, yang2020fda, vu2019advent}, person re-identification~\cite{fu2019self, liu2019adaptive, zeng2020hierarchical}, object detection~\cite{guan2021uncertainty, chen2018domain, xu2020exploring, Meta-DETR, PNPDet}, etc. However, existing UDA methods often suffer from a clear overfitting problem while applied to the scene text detection task. Take the prevalent self-training ~\cite{wu2020synthetic, grandvalet2005semi,zou2018unsupervised,zou2019confidence} as an example. It tackles UDA by pseudo-labeling target samples for network retraining and has demonstrated superb performance while handling multi-class data as in general semantic segmentation and object detection tasks. Scene text detection instead involves a bi-class pseudo-label prediction task (i.e., foreground text and background stuff) where the prediction scores often have an extreme bimodal pattern with two sharp peaks at probabilities 0 and 1 as illustrated in Fig.~\ref{fig:distrib}. As a result, the predictions tend to be over-confidently wrong which introduces clear overfitting together with very high model likelihood and severe noises in pseudo-labelling.

We design SCAST, a subcategory-aware self-training technique that introduces subcategorization for robust UDA for the scene text detection task. For the labelled source-data, SCAST identifies multiple pseudo subcategories for both foreground texts and background stuff via clustering which converts a bi-class prediction task to a multi-class prediction task, leading to more generalizable source models via regularization with a multi-class learning objective. For unlabelled target data, SCAST predicts pseudo labels with both bi-class and subcategory-aware classifiers which mitigates overfitting and allows transferring more diverse and informative source knowledge to the target domain. In addition, it exploits the prediction consistency between the bi-class and subcategory-aware classifiers over target samples which leads to better domain adaptation and more accurate pseudo-labelling of target images.

The contributions of this work are threefold. \textit{First}, we identify the overfitting issue in domain adaptive scene text detection and propose a subcategorization method that mitigates the overfitting effectively. To the best of our knowledge, this is the first work that explores subcategorization in domain adaptive scene text detection tasks. \textit{Second}, we design SCAST that can learn more generalizable source models by identifying subcategories for model regularization. SCAST can transfer more diverse source knowledge and improve pseudo-labeling of target samples with prediction consistency between the bi-class and subcategory-aware classifiers learnt from source data. \textit{Third}, extensive experiments show that SCAST achieve superior detection performance and the subcategorization idea can generalize to other bi-class detection tasks with little adaptation.

\begin{figure*}[t]
    \centering
    \includegraphics[width=\linewidth]{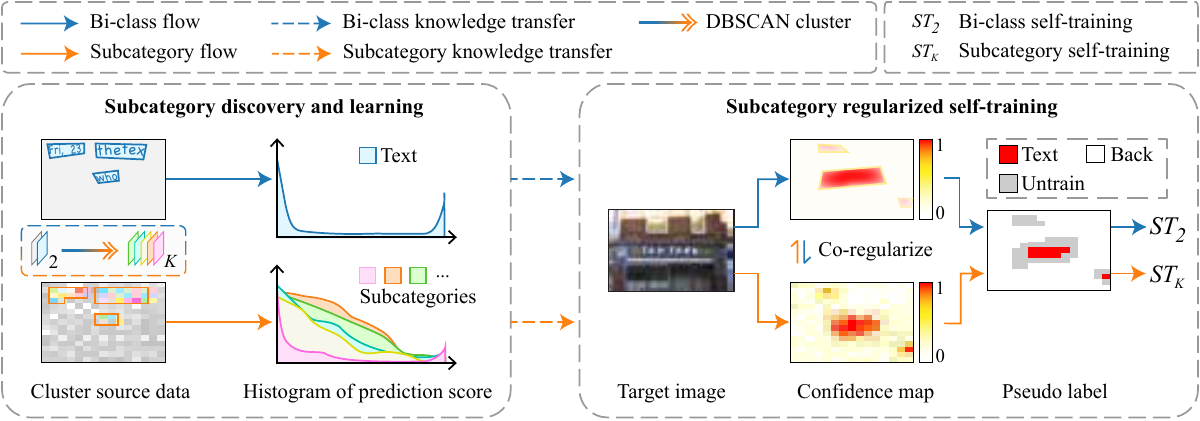}
    \caption{\textbf{The framework of the proposed SCAST:} SCAST mitigates the overfitting of text detectors via subcategorization in the source and the target domains: 1) In the labeled source domain, we conduct feature clustering to discover fine-grained subcategories learning with which relieves the overfitting in bi-class classification greatly; 2) In the unlabeled target domain, we co-regularize pseudo-labelling with the bi-class and multi-class classifiers learnt from source samples, which reduces pseudo labelling noises in self-training effectively.
    }
    \label{fig:overview}
\end{figure*}

\section{Related Work}

\textbf{Scene text detection} aims to locate texts in scene images and it has been widely explored via box regression and image segmentation. The regression approach exploits different object detection frameworks and text-specific shapes and orientations~\cite{zhou2017east, liao2018textboxes++, liao2020mask, lyu2018multi, zhang2019look, tian2017wetext, xue2022detection, xue2019msr, xue2018accurate, xue2022language}. For example, RRD~\cite{liao2018rotation} extracts rotation-sensitive features with SSD~\cite{liu2016ssd} for rotation-invariant detection. Textboxes++~\cite{liao2018textboxes++} modifies convolutional kernels and anchor boxes to capture various text shapes effectively. EAST~\cite{zhou2017east} directly infers quadrangles without proposal mechanism. However, the regression approach often struggle while handling scene texts with irregular shapes. The segmentation approach addresses this issue effectively by predicting a semantic label for each image pixel~\cite{deng2018pixellink, wang2019shape, dai2021progressive, tian2019learning, liao2020real}. For example, PixelLink~\cite{deng2018pixellink} locates text regions by associating neighborhood image pixels. PSENet~\cite{wang2019shape} generates text bounding boxes from multi-scale segmentation maps. DB~\cite{liao2020real} introduces differentiable binarization for adaptive thresholding while predicting bounding regions from segmentation maps.

Many existing methods can achieve very impressive detection performance under the presence of large-scale labelled training images that are often laborious to collect. Domain adaptive scene text detection, which aims to exploit previous collection of annotated scene text images for handling various new data, is largely neglected. Several recent studies~\cite{chen2019cross, wu2020synthetic, chen2021self, zeng2020hierarchical} attempt to tackle this challenge but they focus on adapting existing UDA methods without addressing text-specific problems.

\textbf{Unsupervised domain adaptation (UDA)} has been studied extensively via three typical approaches, namely, adversarial learning, image translation, and self-training. Adversarial learning aligns source and target domains by minimizing certain distribution discrepancy in feature spaces~\cite{bousmalis2016domain, long2015learning, long2017deep, ganin2015unsupervised, ghifary2016deep}. Image translation works by translating source images to have similar target styles~\cite{taigman2016unsupervised, hoffman2018cycada, bousmalis2017unsupervised}. The two approaches cannot handle scene text detection well as they focus on adapting low-frequency appearance information. Self-training works by pseudo-labelling target data, which has been widely explored for various multi-class recognition tasks such as sign recognition~\cite{saito2017asymmetric}, semantic segmentation~\cite{zou2018unsupervised, zou2019confidence} and panoptic segmentation~\cite{huang2021cross, liu2020unsupervised}.

We adopt self-training that pseudo-labels target images with a source-trained model. However, we focus on a more challenging bi-class self-training task that suffers from severe overfitting.  We address the challenge by introducing subcategory pseudo-labeling that mitigates the overfitting by learning from multiple subcategories beyond the original background and foreground texts.

\section{Method}
\label{sec:method}

The proposed subcategory-aware self-training explores subcategorization to mitigate the overfitting in domain adaptive scene text detection task as illustrated in Fig.~\ref{fig:overview}. In the source domain, it aims for \textit{Subcategory Discovery and Learning} that first performs feature clustering to determine multiple subcategories beyond the original two classes and then employs the subcategory labels as pseudo labels to learn multi-class classification model. In the target domain, it aims for \textit{Subcategory Regularized Self-Training} that employs the learnt multi-class classifier to label target samples and retrain the network iteratively, more details to be described in the ensuing subsections.

\subsection{Problem Definition}
\label{sec:prob-def}

Given scene text images in source and target domains $\left\{X^s, X^t\right\}\subset \mathbb{R}^{H\times W\times 3}$ where only the source data $X^s$ are labelled, we aim to learn a scene text detection model that performs well over the target samples $X^t$. The baseline model $\mathbf{G}$ is trained on labeled source data with loss:
\begin{equation}
\label{eq:supervised}
    \mathcal{L}_\mathrm{bi}\left(X^s, Y^s; \mathbf{G}=\mathbf{C}(\mathbf{E})\right),
\end{equation}
where $Y^s$ is bi-class annotations of the source data and $\mathcal{L}_\mathrm{bi}$ denotes a binary classification loss (\textit{e.g.} dice loss~\cite{wang2019shape} or binary cross-entropy loss~\cite{liao2020real, zhou2017east}). The baseline model $\mathbf{G}$ consists of a bi-class classifier $\mathbf{C}$ and a feature extractor $\mathbf{E}$.

\subsection{Subcategory Discovery and Learning}
\label{sec:sub-dl}
In the source domain, we perform feature clustering to discover multiple subcategories on top of the original two-class labels to address the overfitting problem. 
Besides a bi-class classifier trained with the original scene text bounding boxes, a multi-class classifier is trained by using the discovered subcategories as pseudo labels as illustrated in Fig.~\ref{fig:overview}. The multi-class classifier mitigates the overconfident predictions effectively which can be observed in the \textit{Histograms of Prediction Scores} that are produced by the bi-class and multi-class classifiers. 

\textbf{Subcategory Discovery in Source Domain.} 
SCAST discovers subcategories by clustering features instead of spatial relationships among image patches or pixels as in~\cite{hartigan1979algorithm,ester1996density,yue2019domain,tsai2019domain,huang2021mlan,chen2017s}. Feature clustering can produce fine-grained sub-clusters of scene texts that are distinguishable by their high-level features such as textures, strokes, etc.

Given $X^s$, we first obtain their feature maps $f^s$ with a pre-trained feature extractor $\mathbf{E}$ of the baseline model $\mathbf{G}$. DBSCAN~\cite{ester1996density} clustering is then applied to $f^s$ to discover sub-clusters for texts and image background (obtained with the source annotations) separately. The indexes of the identified sub-clusters are then used as pseudo labels $\hat{Y}^s_\mathrm{sub}$ to train a multi-class classifier. The subcategory pseudo-labeling process can be formulated by:
\begin{equation}
    \hat{Y}_\mathrm{sub}^s=\mathbf{\Gamma}\left(f^s\right), f^s=\mathbf{E}\left(X^s\right),
    \label{eq:pseudolabel}
\end{equation}
where $\mathbf{\Gamma}$ denotes the subcategory pseudo labelling operation. For efficiency, we perform clustering on down-sampled feature maps with a down-sample ratio $d$, where each feature point corresponds to a text-height area. 

\begin{figure*}[!t]
    \centering
    \includegraphics[width=\linewidth]{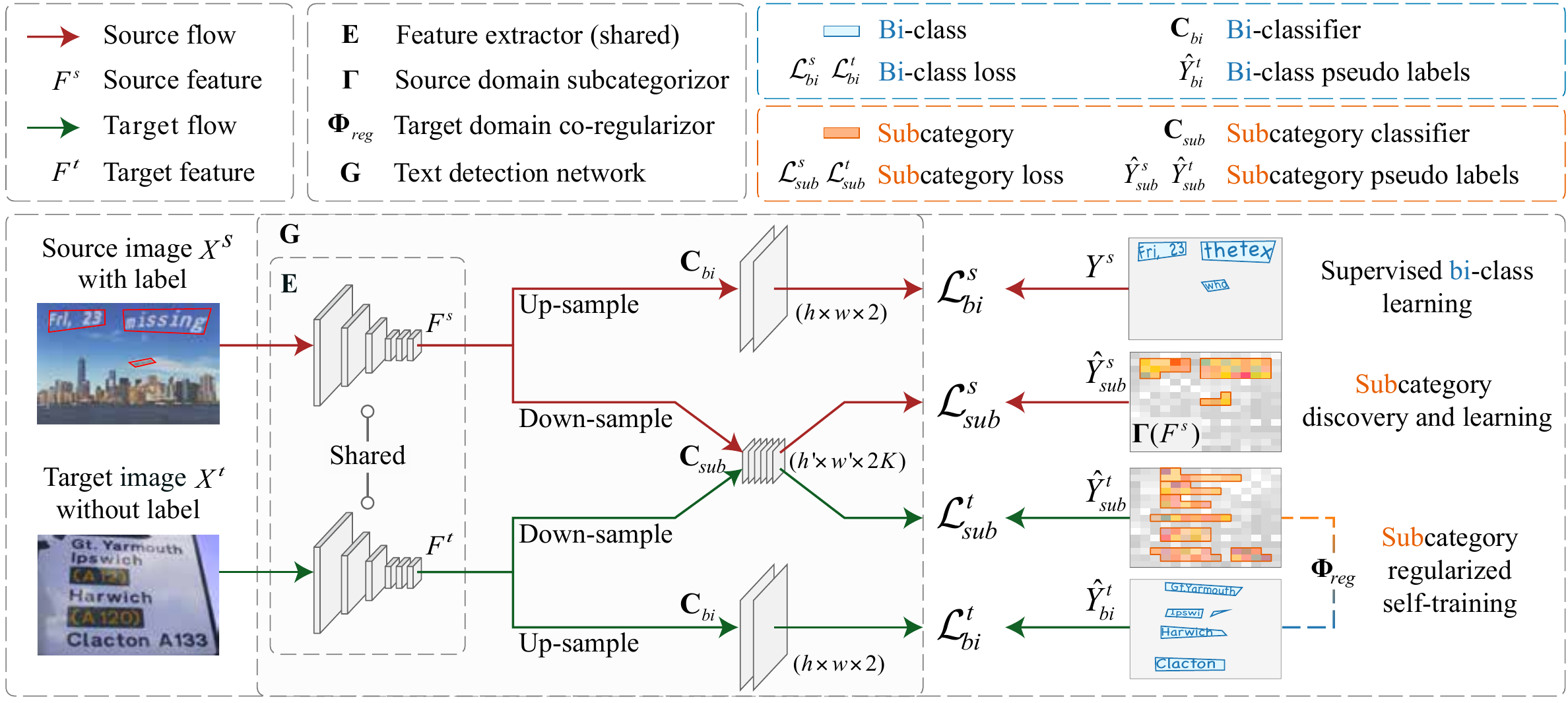}
    \caption{
    \textbf{The framework of the proposed SCAST:} SCAST consists of a multi-class subcategory classifier $\mathbf{C}_\mathrm{sub}$, a source-domain subcategorizer $\mathbf{\Gamma}$, and a target-domain co-regularizer $\mathbf{\Phi}_\mathrm{reg}$. In the source domain, $\mathbf{\Gamma}(\cdot)$ clusters features of source samples to produce multi-class subcategory pseudo labels $\hat{Y}_\mathrm{sub}^s$ and applies them to re-train a network model that suffers from much less overfitting than that trained with the original bi-class labels $Y^s$. In the target domain, $\mathbf{\Phi}_\mathrm{reg}(\cdot)$ co-regularizes between the bi-class and multi-class subcategory pseudo labels $\hat{Y}_\mathrm{bi}^t$ and $\hat{Y}_\mathrm{sub}^t$ to filter out noisy pseudo labels which further mitigates overfitting and improves domain adaptation effectively.
    }
    \label{fig:arch}
\end{figure*}

\textbf{Subcategory Learning in Source Domain.} To learn from the pseudo-labeled subcategories in the source domain, we include a multi-class classifier $\mathbf{C}_\mathrm{sub}$ on top of the feature extractor $\mathbf{E}$. The multi-class subcategory classifier is learnt with a cross-entropy loss:
\begin{multline}
\label{eq:sub_src}    \mathcal{L}_\mathrm{sub}\left(P^s_\mathrm{sub},\hat{Y}^s_\mathrm{sub}\right)=\\-\sum_{h,w}\sum_{k\in K}\left(\hat{Y}_\mathrm{sub}^s\right)^{(h,w,k)} \log \left(\left(P^s_\mathrm{sub}\right)^{(h,w,k)}\right),
\end{multline}
where $P^s_\mathrm{sub}$ is predictions of multi-class classifier and $\hat{Y}_\mathrm{sub}^s$ refers to multi-class pseudo labels (Eq.~\ref{eq:pseudolabel}). We denote the subcategory class number as $K$, which is decided automatically by DBSCAN algorithm. The learned model $\mathbf{G}^s$ is then applied to target domain through a subcategory regularized self-training, introduced next.

\subsection{Subcategory Regularized Self-Training}
\label{sec:sub-reg}
In the target domain, we employ bi-class and subcategory classifiers to predict pseudo labels and co-regularize them for effective knowledge transfer. As illustrated in Fig.~\ref{fig:overview}, the subcategory predictions with less overfitting co-regularize the bi-class predictions, which help filter out noisy pseudo labels as `Untrain' effectively.

\textbf{Subcategory learning in target domain.} We select the most confident predictions by $\mathbf{G}^s$ as the pseudo labels for target images.
\begin{algorithm}[!t]
\caption{Determination of threshold $\theta$}
\label{alg:thetacalc}
\begin{algorithmic}[1]
    \REQUIRE Predictions $P_\mathrm{sub}^t$; Selection portion $\rho\%$
    \FOR {$k=1$ \TO $K$}
        \STATE {$M^k$ = sort($(P^t_\mathrm{sub})^k$, order = descending)}
        \STATE {$l^k$ = length$(M^k)\times \rho\%$}
        \STATE {$\theta^k$ = $M^k[l^k]$}
    \ENDFOR
    \RETURN $\theta=\{\theta^{k}|k\in K\}$
    \end{algorithmic}
\end{algorithm}
As summarized in Algorithm.~\ref{alg:thetacalc}, for each subcategory, we employ a selection proportion $\rho$~\cite{zou2018unsupervised,zou2019confidence} to select the top $\rho\%$ most confident predictions as pseudo labels. With the threshold value $\theta$ at the top $\rho\%$-th position, the pseudo-labeling process can be formulated as:
\begin{equation}
\label{eq:thres}
    \hat{Y}^t_\mathrm{sub}=\mathbf{\Theta}\left(\theta, P^t_\mathrm{sub}\right)=
    \begin{cases}
      1, & \text{if}\ (P^t_\mathrm{sub})^{k}>\theta^{k},\\
      0, & \text{otherwise},\\
    \end{cases}
\end{equation}
where $k\in K$ is the subcategory index, and $\hat{Y}^t_\mathrm{sub}$ is the subcategory pseudo labels. The bi-class pseudo labelling follows the same formula: $\hat{Y}^t_\mathrm{bi}=\mathbf{\Theta}\left(\theta, P^t_\mathrm{bi}\right)$ where $k=0,1$ refers to texts or image background, respectively.

The network can then be re-trained with the target data with the obtained pseudo labels $\left\{\hat{Y}^t_\mathrm{bi}, \hat{Y}^t_\mathrm{sub}\right\}$. The training losses are the same as the supervised losses $\mathcal{L}_\mathrm{bi}$ and $\mathcal{L}_\mathrm{sub}$ (in Eq.~\ref{eq:supervised} and Eq.~\ref{eq:sub_src}) in the source domain. The learning objective in the target domain is thus a weighted combination of bi-class loss and subcategory loss:
\begin{align}
\label{eq:pslearn}
    \begin{split}
        \mathcal{L}^t = & \lambda_\mathrm{sub}\mathcal{L}_\mathrm{sub}\left(P_\mathrm{sub}^t,\hat{Y}_\mathrm{sub}^t;\mathbf{C}_\mathrm{sub}, \mathbf{E}\right) \ +\\
        & \lambda_\mathrm{bi}\mathcal{L}_\mathrm{bi}\left(P_\mathrm{bi}^t,\hat{Y}_\mathrm{bi}^t;\mathbf{C}_\mathrm{bi}, \mathbf{E}\right).
    \end{split}
\end{align}
where $\lambda_\mathrm{bi}$ is a weight parameter and we set it empirically at 1 in our implementation. Eq.~\ref{eq:thres} and Eq.~\ref{eq:pslearn} define a single iteration of self-training that generates pseudo labels and retrains the network in single-round. In practice, we implement multi-round optimization as described in Sec~\ref{sec:net-opt}.

\textbf{Subcategory regularization in target domain.} As shown in Fig.~\ref{fig:overview}, we discard inconsistent predictions between the bi-class and subcategory classifiers, which effectively reduces false positives and pseudo-label noises and leads to more robust pseudo-labeling.
Specifically, we measure the distance between bi-class predictions and subcategory predictions by cross-entropy and drop out predictions with large distances (Eq.~\ref{eq:reg-selection}).
To control the proportion of dropped predictions, we set a ratio $\rho_\mathrm{reg}$ to drop out positions with top $\rho_\mathrm{reg}\%$ distance (we set $\rho_\mathrm{reg}\%=10\%$ empirically).
We denote such co-regularization process as $\mathbf{\Phi}_\mathrm{reg}$ and formulate it as a loss minimization problem:

\begin{align}
\label{eq:ce-reg}
    \mathbf{\Phi}_\mathrm{reg}=\argmin_{\left\{P^t_\mathrm{bi}, P^t_\mathrm{sub}\right\}} \mathcal{L}^t_\mathrm{reg}\left(P^t_\mathrm{bi}, P^t_\mathrm{sub}; \rho_\mathrm{reg}\right),
\end{align}
\noindent where $\mathcal{L}^t_\mathrm{reg}\left(P^t_\mathrm{bi}, P^t_\mathrm{sub}\right)=-\left(P^t_\mathrm{sub}\right) \log \left(P^t_\mathrm{bi}\right)$ is the cross-entropy distance between bi- and multi-class predictions. We can get regularized pseudo labels by solving Eq.~\ref{eq:ce-reg}:
\begin{equation}
\label{eq:reg-selection}
    \left\{\hat{Y}^t_\mathrm{bi}, \hat{Y}^t_\mathrm{sub}\right\} = 
    \begin{cases}
        1, & \text{if}\ \mathcal{L}^t_\mathrm{reg}\left(P^t_\mathrm{bi}, P^t_\mathrm{sub}\right)<\theta_\mathrm{reg},\\
        0, & \text{otherwise},\\
    \end{cases}
\end{equation}
where $\theta_\mathrm{reg}$ is the threshold value at top $\rho_\mathrm{reg}\%$-th position.

\subsection{Network Optimization}
\label{sec:net-opt}

\begin{table*}[t]
\centering
\resizebox{\textwidth}{!}{%
\begin{tabular}{@{}ccccccccccccccccc@{}}
\toprule
\multirow{2}{*}{Method} & \multirow{2}{*}{Arch.} & \multicolumn{3}{c}{ICDAR13~\cite{karatzas2013icdar}} & \multicolumn{3}{c}{ICDAR15~\cite{karatzas2015icdar}} & \multicolumn{3}{c}{COCO-Text~\cite{veit2016coco}} & \multicolumn{3}{c}{Total-Text (Reg)~\cite{shi2017icdar2017}} & \multicolumn{3}{c}{Mean} \\ \cmidrule(l){3-17} 
 &  & Precision & Recall & F-Score & Precision & Recall & F-Score & Precision & Recall & F-Score & Precision & Recall & F-Score & Precision & Recall & F-Score \\ \midrule
\textit{Supervised} & - & 92.64 & 82.67 & 87.37 & 84.36 & 81.27 & 82.79 & 50.39 & 32.40 & 39.45 & 50.00 & 36.20 & 42.00 & 69.35 & 58.14 & 62.90 \\ \midrule
Baseline (EAST~\cite{zhou2017east}) & - & 65.33 & 68.85 & 67.05 & 69.63 & 53.44 & 60.47 & 38.23 & 20.73 & 26.89 & 40.35 & 36.61 & 38.39 & 53.38 & 44.90 & 48.20 \\
TST~\cite{wu2020synthetic} & \textit{ST} & 71.50 & 70.70 & 71.10 & 69.30 & 60.50 & 64.60 & 53.00 & 22.76 & 31.85 & 43.48 & 37.56 & 40.30 & 59.32 & 47.88 & 51.96 \\
EntMin~\cite{grandvalet2005semi} & \textit{ST} & 68.29 & 67.81 & 68.05 & 70.41 & 53.16 & 60.58 & \textbf{57.12} & 18.52 & 27.97 & 43.99 & 34.50 & 38.67 & 59.95 & 43.50 & 48.82 \\
CBST~\cite{zou2018unsupervised} & \textit{ST} & 70.28 & 71.08 & 70.68 & 71.36 & 58.40 & 64.23 & 55.17 & 20.96 & 30.38 & 43.17 & 38.23 & 40.55 & 60.00 & 47.17 & 51.46 \\
CRST~\cite{zou2019confidence} & \textit{ST} & 72.21 & 73.05 & 72.63 & 71.85 & 62.19 & 66.67 & 51.44 & 22.28 & 31.09 & 46.31 & 39.13 & 42.42 & 60.45 & 49.16 & 53.20 \\
FDA~\cite{yang2020fda} & \textit{Tran.} & 47.28 & 73.88 & 57.66 & 59.13 & 55.18 & 57.09 & 44.69 & 19.79 & 27.43 & 37.24 & 40.15 & 38.64 & 43.86 & 47.02 & 43.35 \\
ADVENT~\cite{vu2019advent} & \textit{Adv.} & 62.15 & 66.30 & 64.16 & 63.00 & 61.00 & 61.99 & 38.60 & 21.06 & 27.25 & 38.44 & 37.51 & 37.97 & 50.55 & 46.47 & 47.84 \\ \midrule
Ours (SCAST) & \textit{ST} & \textbf{79.88} & \textbf{76.65} & \textbf{78.23} & \textbf{75.80} & \textbf{65.27} & \textbf{70.14} & 49.28 & \textbf{26.07} & \textbf{34.10} & \textbf{52.86} & \textbf{42.97} & \textbf{47.40} & \textbf{64.46} & \textbf{52.74} & \textbf{57.47} \\ \bottomrule
\end{tabular}
}
\caption{\textbf{Regular scene text detection:} The experiments are conducted over domain adaptive scene text detection tasks SynthText $\to$ \{ICDAR13, ICDAR15, COCO-Text and Total-Text~(Reg)\} using EAST~\cite{zhou2017east}. \textit{ST} stands for Self-training, \textit{Adv.} for Adversarial training, and \textit{Tran.} for Image-to image translation. \textit{Supervised} refers to supervised learning from the labeled target data.
}
\label{tab:reg-sota}
\end{table*}

\begin{table*}[t]
\centering
\resizebox{\textwidth}{!}{%
\begin{tabular}{@{}cccccccccccccc@{}}
\toprule
\multirow{2}{*}{Method} & \multirow{2}{*}{Arch.} & \multicolumn{3}{c}{Total-Text (Curve)~\cite{shi2017icdar2017}} & \multicolumn{3}{c}{CTW1500~\cite{liu2019curved}} & \multicolumn{3}{c}{TextSeg~\cite{xu2021rethinking}} & \multicolumn{3}{c}{Mean} \\ \cmidrule(l){3-14} 
 &  & Precision & Recall & F-Score & Precision & Recall & F-Score & Precision & Recall & F-Score & Precision & Recall & F-Score \\ \midrule
\textit{Supervised} & - & 81.77 & 75.11 & 78.30 & 80.57 & 75.55 & 78.00 & 89.02 & 81.55 & 85.12 & 83.79 & 77.40 & 80.47 \\ \midrule
Baseline (PSENet~\cite{wang2019shape}) & - & 72.41 & 46.12 & 56.35 & 44.03 & 44.26 & 44.15 & 77.24 & 49.33 & 60.20 & 64.56 & 46.57 & 53.57 \\
TST~\cite{wu2020synthetic} & \textit{ST} & 68.94 & 51.03 & 58.65 & 42.07 & 50.98 & 46.09 & 82.91 & 51.03 & 63.17 & 64.64 & 51.01 & 55.97 \\
EntMin~\cite{grandvalet2005semi} & \textit{ST} & 72.20 & 46.33 & 56.44 & 43.38 & 48.50 & 45.80 & 82.23 & 48.34 & 60.88 & 65.94 & 47.72 & 54.37 \\
CBST~\cite{zou2018unsupervised} & \textit{ST} & 71.39 & 47.57 & 57.09 & 44.58 & 49.19 & 46.77 & 79.17 & 52.15 & 62.88 & 65.05 & 49.64 & 55.58 \\
CRST~\cite{zou2019confidence} & \textit{ST} & 73.64 & 50.31 & 59.78 & 46.35 & 47.75 & 47.04 & 82.30 & 51.41 & 63.29 & 67.43 & 49.82 & 56.70 \\
FDA~\cite{yang2020fda} & \textit{Tran.} & 71.60 & 43.50 & 54.12 & 45.02 & 44.38 & 44.70 & 78.21 & 50.02 & 61.02 & 64.94 & 45.97 & 53.28 \\
ADVENT~\cite{vu2019advent} & \textit{Adv.} & 76.27 & 34.34 & 47.36 & 41.37 & 46.97 & 43.99 & 80.83 & 50.04 & 61.81 & 66.16 & 43.78 & 51.05 \\ \midrule
Ours (SCAST) & \textit{ST} & \textbf{71.82} & \textbf{56.73} & \textbf{63.39} & \textbf{51.02} & \textbf{56.10} & \textbf{53.44} & \textbf{81.90} & \textbf{58.78} & \textbf{68.44} & \textbf{68.25} & \textbf{57.20} & \textbf{61.76} \\ \bottomrule
\end{tabular}
}
\caption{\textbf{Irregular scene text detection: }The experiments are conducted over domain adaptive scene text detection tasks SynthText $\to$ \{Total-Text (Curve), CTW1500, TextSeg\} using PSENet~\cite{wang2019shape}. \textit{ST} stands for Self-training, \textit{Adv.} for Adversarial training, and \textit{Tran.} for Image translation architecture.
\textit{Supervised} refers to supervised learning from the labeled target data.
}
\label{tab:irreg-sota}
\end{table*}

\begin{figure*}[t]
    \centering
    \includegraphics[width=\linewidth]{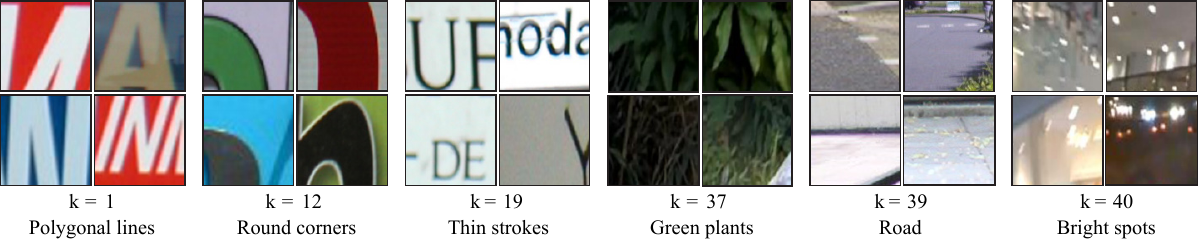}
    \caption{\textbf{Samples of clustered subcategories:} Our method captures high-level features ($e.g.$ curvature, texture, stroke shapes, hue, etc.) and clusters text instances and background into different subcategories with different features. The experiments were conducted on ICDAR13 with EAST detector. $k$ is the subcategory index and we show five subcategories ($k=1$, 12 and 19 for text, 37 and 40 for background) for illustration.
    }
    \label{fig:illsub}
\end{figure*}

\begin{table*}[t]
\centering
\resizebox{\textwidth}{!}{%
\begin{tabular}{@{}cccccccccccccc@{}}
\toprule
\multirow{2}{*}{Method} & \multirow{2}{*}{Arch.} & \multicolumn{3}{c}{ICDAR13~\cite{karatzas2013icdar}} & \multicolumn{3}{c}{ICDAR15~\cite{veit2016coco}} & \multicolumn{3}{c}{Total-Text (Reg)~\cite{shi2017icdar2017}} & \multicolumn{3}{c}{Mean} \\ \cmidrule(l){3-14} 
 &  & Precision & Recall & F-Score & Precision & Recall & F-Score & Precision & Recall & F-Score & Precision & Recall & F-Score \\ \midrule
\textit{Supervised} & - & 92.64 & 82.67 & 87.37 & 84.36 & 81.27 & 82.79 & 50.00 & 36.20 & 42.00 & 75.67 & 66.71 & 70.72 \\ \midrule
Baseline (EAST~\cite{zhou2017east}) & - & 62.31 & 70.31 & 66.07 & 71.29 & 55.20 & 62.22 & 41.03 & 35.28 & 37.94 & 58.21 & 53.60 & 55.41 \\
TST~\cite{wu2020synthetic} & \textit{ST} & 66.27 & 72.05 & 69.04 & 73.46 & 57.21 & 64.32 & 42.23 & 38.10 & 40.06 & 60.65 & 55.79 & 57.81 \\
EntMin~\cite{grandvalet2005semi} & \textit{ST} & 64.13 & 72.98 & 68.27 & 70.68 & 56.08 & 62.54 & 44.06 & 36.19 & 39.74 & 59.62 & 55.08 & 56.85 \\
CBST~\cite{zou2018unsupervised} & \textit{ST} & 68.68 & 71.36 & 69.99 & 69.05 & 59.30 & 63.80 & 42.28 & 37.92 & 39.98 & 60.00 & 56.19 & 57.93 \\
CRST~\cite{zou2019confidence} & \textit{ST} & 71.33 & 74.42 & 72.84 & 73.89 & 63.09 & 68.06 & 43.60 & 39.33 & 41.36 & 62.94 & 58.95 & 60.75 \\
FDA~\cite{yang2020fda} & \textit{Tran.} & 43.62 & 61.50 & 51.04 & 62.07 & 52.75 & 57.03 & 37.92 & 33.40 & 35.52 & 47.87 & 49.22 & 47.86 \\
ADVENT~\cite{vu2019advent} & \textit{Adv.} & 57.42 & 61.96 & 59.60 & 65.83 & 51.47 & 57.77 & 43.15 & 33.29 & 37.58 & 55.47 & 48.91 & 51.65 \\ \midrule
Ours (SCAST) & \textit{ST} & \textbf{76.48} & \textbf{78.26} & \textbf{77.36} & \textbf{77.24} & \textbf{67.40} & \textbf{71.99} & \textbf{47.06} & \textbf{41.75} & \textbf{44.25} & \textbf{66.93} & \textbf{62.47} & \textbf{64.53} \\ \bottomrule
\end{tabular}
}
\caption{\textbf{Real to real adaptation:} The experiments are conducted over real-to-real domain adaptive scene text detection tasks COCO-Text $\to$ \{ICDAR13, ICDAR15 and Total-Text~(Reg)\} using EAST~\cite{zhou2017east}. \textit{ST} stands for Self-training, \textit{Adv.} for Adversarial training, and \textit{Tran.} for Image translation architecture. \textit{Supervised} refers to supervised learning from the labeled target data.}
\label{tab:real-real}
\end{table*}

We optimize SCAST by multiple rounds of self-training. Each round consists of two alternate steps including updating of pseudo labels $\{\hat{Y}^t_\mathrm{bi}, \hat{Y}^t_\mathrm{sub}\}$ and retraining network $\mathbf{G}$ with the updated pseudo labels. The optimization can be formulated as a unified loss minimization problem:
\begin{equation}
    \argmin_{\mathbf{G},\hat{Y}^t_\mathrm{bi},\hat{Y}^t_\mathrm{sub}}\mathcal{L}^t\left(P^t_\mathrm{bi}, P^t_\mathrm{sub}, \hat{Y}^t_\mathrm{bi}, \hat{Y}^t_\mathrm{sub};\mathbf{G}, \theta\right), \label{eq:net_opt}
\end{equation}
where $\mathbf{G}(\cdot)=\left\{\mathbf{C}_\mathrm{bi}\left(\mathbf{E}(\cdot)\right),\mathbf{C}_\mathrm{sub}\left(\mathbf{E}(\cdot)\right)\right\}$ and $\mathcal{L}^t$ is the weighted cross-entropy loss (Eq.~\ref{eq:pslearn}).

\textbf{Pseudo label prediction.} We fix network $\mathbf{G}$ while predicting pseudo labels $\hat{Y}^t_\mathrm{bi}$ and $\hat{Y}^t_\mathrm{sub}$ with thresholds $\theta$ in each round. We predict pseudo labels in an ``easy-to-hard" manner so that the network can learn from confident pseudo labels first. This strategy helps reduce the impact of noisy pseudo labels for both bi-class multi-class objectives effectively. In implementation, we increase the selection proportion $\rho\%$ gradually after each training round which reduces the corresponding thresholds $\theta$ accordingly. The optimization in this step can be formulated by:
\begin{equation}
    \argmin_{\hat{Y}_\mathrm{sub}^t, \hat{Y}_\mathrm{bi}^t}\mathcal{L}^t\left(P^t_\mathrm{bi}, P^t_\mathrm{sub}, \hat{Y}^t_\mathrm{bi}, \hat{Y}^t_\mathrm{sub};\mathbf{G}, \theta_{(i)}\right), \label{eq:label_update}
\end{equation}
where $i$ is the optimization round.

\textbf{Update the network.} We fix pseudo labels while retraining $\mathbf{G}$. As more pseudo labels are selected round by round, model $\mathbf{G}$ learns and mitigates the domain gap gradually. The optimization can be formulated by:
\begin{equation}
    \argmin_\mathbf{G}\mathcal{L}^t\left(P^t_\mathrm{bi}, P^t_\mathrm{sub}, \hat{Y}^t_\mathrm{bi}, \hat{Y}^t_\mathrm{sub};\mathbf{G}, \theta_{(i)}\right), \label{eq:net_update}
\end{equation}
where $i$ is the optimization round.

\section{Experiments}

\subsection{Datasets and Evaluation}
\textbf{Datasets.} Our experiments involve several domain adaptive scene text detection tasks including SynthText $\to$ \{ICDAR13, ICDAR15, COCO-Text17, Total-Text~(Reg)\} for regular scene text and SynthText $\to$ \{Total-Text~(Curve), CTW1500, TextSeg\} for irregular scene text. More dataset details are provided in the appendix.

\textbf{Evaluation.} We evaluate using Precision, Recall and F-Score as in ~\cite{karatzas2015icdar,karatzas2013icdar}. The evaluations are based on the Intersection-over-Union (IoU) criterion in PASCAL~\cite{everingham2015pascal}, with a widely adopted threshold of $50\%$. During training and evaluations, we ignore unreadable text regions that are labeled by either ``do not care" or ``illegal" in all datasets. Following previous UDA methods~\cite{zou2018unsupervised, zou2019confidence, vu2019advent}, we adopt dense evaluation in the experiments.
\vspace{-0.4em}

\subsection{Implementation Details}
\vspace{-0.4em}
\textbf{Subcategorizor.} We perform DBSCAN clustering on feature maps to discover subcategories. Specifically, we down-sample the feature maps to make each feature point correspond to a text-height region. We conduct DBSCAN clustering on these feature maps for both text and background (the distance $\epsilon$ in DBSCAN is discussed in Sec.~\ref{sec:abl}).

\textbf{Network architectures.} We evaluate our method with regular text detector EAST~\cite{zhou2017east} and irregular text detector PSENet~\cite{wang2019shape}. The two detectors have simple detection strategies without complex post-processing, which reduce interference in evaluations. For EAST, we adopt the settings of the officially released code, which use VGG-16 pre-trained model on FPN~\cite{lin2017feature} as the backbone, and dice-loss~\cite{sudre2017generalised} as classification loss. During post-processing, we use Local-Aware NMS with a threshold $0.5$, same as the original paper. For PSENet, we follow PSENet-1s from the original paper, which uses ResNet~\cite{he2016deep} pre-trained on ImageNet~\cite{deng2009imagenet} as backbone. In addition, we train the network from scratch on SynthText without pretraining on the ICDAR17-MLT~\cite{nayef2017icdar2017} to avoid introducing extra data.

\textbf{Training details.} We use SGD optimizer with momentum $0.9$ and a weight decay $5e-4$ in training. The initial learning rate is $1e-3$ and then decays by a polynomial policy of power $0.9$. For training with source data, we adopted all data augmentation strategies as used in the original papers. For training with target data, we remove all data augmentations and resize training images to $512\times512$. We set the training batch-size at $12$, including $6$ source images and $6$ target images during domain adaption.

\textbf{Self-Training details.} The pseudo label proportion $\rho\%$ and related threshold $\theta$ are important for pseudo label prediction. To avoid heuristic setups, we adopt linearly increased $\rho\in\{20,40,60,80,100\}$ to select the top $\{20\%,40\%,60\%,80\%,100\%\}$ confident predictions as pseudo labels.

\subsection{Ablation Study and Analysis}
\label{sec:abl}

We study the contributions of each design in the proposed SCAST, including a $K$-Subcategorizor $SC_\mathrm{K}$ for source data, a subcategory self-training $ST_\mathrm{K}$ and a subcategory regularized self-training $ST_\mathrm{K,reg}$ over the target data. 

As Table~\ref{tab:ablation} shows, including subcategorization in $+SC_\mathrm{K}$ in the source domain outperforms the \textit{Baseline} by a large margin, showing that the proposed subcategorization mitigates the overfitting effectively. For target data, including the bi-class self-training with {$+ST_\mathrm{2}$} improves the scene text detection greatly, and combining the source-domain subcategorization with the target-domain bi-class self-training in $+ST_\mathrm{2} + SC_\mathrm{K}$ further improves the detection by a large margin. In addition, including the multi-class subcategory self-training on target data in $+ST_\mathrm{2} + SC_\mathrm{K} + ST_\mathrm{K}$ also improves clearly. Finally, the inclusion of co-regularization between bi-class and multi-class pseudo labels of target data in $+ST_\mathrm{2} + SC_\mathrm{K} + ST_{K, reg}$ produces the best detection. The ablation studies show the effectiveness of the proposed subcategorization and subcategory regularized self-training.

\subsection{Comparison with SOTA}

\begin{table}[!t]
\centering
\resizebox{0.48\textwidth}{!}{%
\begin{tabular}{@{}lcccccccc@{}}
\toprule
Method & $\mathcal{L}^s_\mathrm{bi}$ & $\mathcal{L}^s_\mathrm{sub}$ & $\mathcal{L}^t_\mathrm{bi}$ & $\mathcal{L}^t_\mathrm{sub}$ & $\mathcal{L}^t_\mathrm{reg}$ & {\ \ Prec.\ } & {\ Rec.\ } & {\ F-sco.\ } \\ \midrule
Baseline & \checkmark &  &  &  &  & 69.63 & 53.44 & 60.47 \\
+ $SC_\mathrm{K}$ & \checkmark & \checkmark &  &  &  & 67.94 & 62.93 & 65.34 \\ \midrule
+ $ST_\mathrm{2}$ & \checkmark &  & \checkmark &  &  & 71.36 & 58.40 & 64.23 \\
+ $ST_\mathrm{2}$ + $SC_\mathrm{K}$ & \checkmark & \checkmark & \checkmark &  &  & 75.12 & 63.01 & 68.53 \\
+ $ST_\mathrm{2}$ + $SC_\mathrm{K}$ + $ST_\mathrm{K}$ & \checkmark & \checkmark & \checkmark & \checkmark &  & 73.09 & 65.98 & 69.35 \\
+ $ST_\mathrm{2}$ + $SC_\mathrm{K}$ + $ST_{K, reg}$ & \checkmark & \checkmark & \checkmark & \checkmark & \checkmark & \textbf{75.80} & \textbf{65.27} & \textbf{70.14} \\ \bottomrule
\end{tabular}
}
\caption{
\textbf{Ablation study} of SCAST over domain adaptive scene text detection task SynthText $\to$ ICDAR15 using EAST. $SC_\mathrm{K}$, $ST_\mathrm{2}$, $ST_\mathrm{K}$, and $ST_{K, reg}$ refer to subcategorization (via $\mathcal{L}^s_\mathrm{sub}$), bi-class CBST~\cite{zou2018unsupervised} (via $\mathcal{L}^t_\mathrm{bi}$), subcategory self-training (via $\mathcal{L}^t_\mathrm{sub}$), and subcategory regularized self-training (via $\mathcal{L}^t_\mathrm{reg}$), respectively.}
\label{tab:ablation}
\end{table}

\begin{table}[t]
\centering
\resizebox{0.48\textwidth}{!}{%
\begin{tabular}{|c|c|cc|c|c|c|}
\hline
\multirow{2}{*}{Method} & \multirow{2}{*}{$\epsilon$} & \multicolumn{2}{c|}{Subcategory Number} & \multirow{2}{*}{\ Prec.\ } & \multirow{2}{*}{\ Rec.\ } & \multirow{2}{*}{\ F-sc.\ } \\ \cline{3-4}
 &  & \multicolumn{1}{c|}{\ \ \ \ \textit{Text}\ \ \ \ \ } & {\textit{Background}} &  &  &  \\ \hline
Baseline & - & \multicolumn{1}{c|}{-} & - & \textbf{69.63} & 53.44 & 60.47 \\ \hline
$+SC_{K=15}$ & $0.5$ & \multicolumn{1}{c|}{6} & 9 & 64.28 & 60.82 & 62.50 \\ \hline
$+SC_{K=21}$ & $0.1$ & \multicolumn{1}{c|}{10} & 11 & 69.48 & 59.30 & 63.99 \\ \hline
$+SC_{K=33}$ & $0.05$ & \multicolumn{1}{c|}{19} & 14 & 65.66 & 62.27 & 63.92 \\ \hline
$+SC_{K=54}$ & $0.01$ & \multicolumn{1}{c|}{35} & 19 & 67.94 & \textbf{62.93} & \textbf{65.34} \\ \hline
\end{tabular}
}
\caption{\textbf{Maximum cluster distance $\epsilon$}. With the decrease of the maximum cluster distance $\epsilon$, DBSCAN clustering discovers more subcategories. $SC_\mathrm{K}$ stands for $K$-classes SubCategorization in the source domain, where $K$ equals to the number of subcategories in both \textit{Text} and \textit{Background}. The experiments were performed over task SynthText $\to$ ICDAR15 with EAST backbone.}
\label{tab:epsilon}
\end{table}

\begin{table*}[t]
\centering
\resizebox{\textwidth}{!}{%
\begin{tabular}{@{}lcccccccccccc@{}}
\toprule
\multirow{2}{*}{Network} & \multicolumn{3}{c}{ICDAR13} & \multicolumn{3}{c}{ICDAR15} & \multicolumn{3}{c}{COCO-Text} & \multicolumn{3}{c}{Total-Text} \\ \cmidrule(l){2-13} 
 & Entropy & Err. Rate & Likelihood & Entropy & Err. Rate & Likelihood & Entropy & Err. Rate & Likelihood & Entropy & Err. Rate & Likelihood \\ \midrule
Baseline (Text) & 0.0092 & 38.49\% & 844.79 & 0.0099 & 63.35\% & 833.52 & 0.0101 & 67.83\% & 832.02 & 0.0091 & 68.49\% & 844.80 \\
+ $SC_\mathrm{K}$ & 0.1553 & 33.16\% & 587.70 & 0.2305 & 58.90\% & 537.10 & 0.2724 & 67.92\% & 511.69 & 0.2222 & 63.65\% & 587.70 \\ \midrule
Baseline (Back) & 0.0007 & 0.78\% & 1101.30 & 0.0001 & 0.15\% & 1262.55 & 0.0002 & 0.21\% & 1101.30 & 0.0006 & 0.83\% & 1116.67 \\
+ $SC_\mathrm{K}$ & 0.1186 & 0.70\% & 653.80 & 0.1013 & 0.11\% & 770.34 & 0.1451 & 0.15\% & 653.80 & 0.1290 & 0.96\% & 648.00 \\ \bottomrule
\end{tabular}
}
\caption{\textbf{Overfitting mitigation.} Experiments are conducted on SynthText $\to$ ICDAR15 with EAST~\cite{zhou2017east}. Baseline (Text) and Baseline (Back) measure overfitting of text and background with the baseline model. $SC_\mathrm{K}$ denotes the proposed $K$-class subcategorization. `Likelihood' is described in page 2 footnote and `Err. Rate' is classification error rate.}
\label{tab:overfit}
\end{table*}

\begin{figure*}[t]
    \centering
    \includegraphics[width=\linewidth]{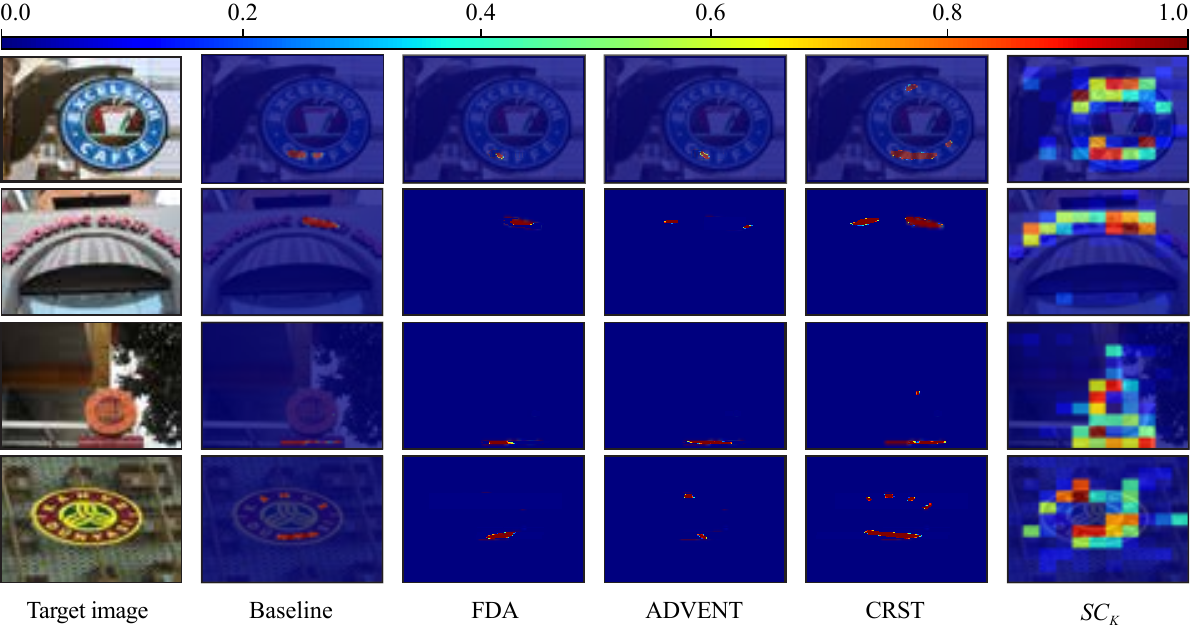}
    \caption{\textbf{Comparison with the state-of-the-art.} The Baseline detector and the compared methods produce over-confident predictions with an extreme bimodal distribution (\textit{i.e.}, most prediction probabilities are around $0$ or $1$), indicating overfitting with degraded detection (with more false negatives). The proposed subcategory-aware detector instead generates prediction probabilities that smoothly distribute between $0$ and $1$, indicating less overfitting. The evaluation is performed on task SynthText $\to$ TotalText with EAST detector.}
    \label{fig:illcurve}
\end{figure*}

We compare SCAST with several state-of-the-art UDA methods that adopt self-training with EntMin~\cite{grandvalet2005semi}, TST~\cite{wu2020synthetic}, CBST~\cite{zou2018unsupervised}, and CRST~\cite{zou2019confidence}), image translation with FDA~\cite{yang2020fda} and adversarial with ADVENT~\cite{vu2019advent}.

Tables~\ref{tab:reg-sota} and ~\ref{tab:irreg-sota} show experimental results on regular and irregular scene text under the synthetic-to-real setup, respectively. We also evaluate real-to-real adaptation to show the robustness of our method in Table~\ref{tab:real-real}. We can see that SCAST outperforms all compared methods in F-score. Specifically, methods using image translation and adversarial learning do not perform well as image translation tends to smooth text strokes while adversarial learning focuses on aligning image background. For UDA using self-training, TST~\cite{wu2020synthetic} and CRST~\cite{zou2019confidence} outperform the Baseline clearly as retraining using pseudo-labeled target data often produces stronger models. However, the improvements are limited as their predicted pseudo labels are noisy due to the overfitted classifiers as shown in Table~\ref{tab:overfit}. As a comparison, SCAST mitigates the overfitting and reduces pseudo label errors and produces clearly better detection as illustrated in Fig.~\ref{fig:illcurve} (more samples available in appendix). Note TST combines adversarial learning with self-training while SCAST performs self-training alone. 

Note SCAST even outperforms supervised models over the Total-Text (Reg) as shown in Tables~\ref{tab:reg-sota} and~\ref{tab:real-real}. The outstanding performance is largely because the subcategory-aware classifier predicts text regions as multiple fragments instead of a single quadrangle. Such predictions are much more compatible for the detection of curved text instances which cannot be located well with quadrangles.

\subsection{Discussion}
\label{sec:discussion}

\textbf{Maximum cluster distance $\epsilon$.} Table~\ref{tab:epsilon} shows the impact of the maximum cluster distance $\epsilon$ in DBSCAN clustering, where the baseline is EAST detector. We can observe a trade-off between the recall and precision, showing that our method greatly improves the recall ($53.44$ to $62.93$) while introducing marginal decrease in precision ($69.63$ to $67.94$) which can be compensated with self-training over target data. We adopt $\epsilon=0.01$ in our implemented system.

\textbf{Subcategorization.} Fig.~\ref{fig:illsub} shows subcategorized foreground texts and image backgrounds. We can see that DBSCAN clustering can group image patches into subcategories with similar features in polygonal lines, traffic signs, curved corners and thin strokes (e.g., subcategories 1, 12, 19, 37, and 40 for illustration). Fig.~\ref{fig:illcurve} further shows subcategory predictions where over-confident predictions are mitigated and False Negatives are reduced effectively. 

\textbf{Overfitting mitigation.} We adopt entropy, error rate, and likelihood metrics to evaluate how the proposed subcategorization helps mitigate the overfitting in domain adaptive scene text detection. Table~\ref{tab:overfit} shows experimental results over four widely adopted datasets. It can be observed that including the proposed subcategorization (i.e., via \textit{+$SC_k$}) helps mitigate the overfitting effectively with higher entropy, fewer pseudo-labeling errors, and lower likelihood.

\textbf{Other bi-class tasks} The proposed SCAST can generalize well to other bi-class detection tasks. We evaluate another two bi-class detection tasks on vehicle detection and pedestrian detection. Experiments over the adaptation task GTA5~\cite{Richter_2016_ECCV} $\rightarrow$ Cityscapes~\cite{Cordts2016Cityscapes} show that SCAST outperforms the state-of-the-art by 2.9\% and 3.1\% (in foreground IoU), respectively. Please refer to the appendix for details.

\section{Conclusion}
This paper presents SCAST, a subcategory-aware self-training technique for mitigating overfitting in domain adaptive scene text detection. We subcategorize source data by feature clustering which mitigates the overfitting substantially. For target data, we design a subcategory regularized self-training that employs the source-learned subcategory priors to co-regularizes bi-class and multi-class subcategory pseudo labels which further mitigates the overfitting effectively. Extensive experiments show that SCAST mitigates the overfitting in scene text detection task with superior detection performance. The idea of subcategorization is generic and can adapt to other bi-class detection and recognition tasks. We will explore it in our future research.

\newpage

%%%%%%%%% REFERENCES
{\small
\bibliographystyle{ieee_fullname}
\bibliography{egbib}
}

\end{document}